\documentclass[11pt]{article}

\usepackage[preprint]{acl}

\usepackage{times}
\usepackage{latexsym}

\usepackage[T1]{fontenc}

\usepackage[utf8]{inputenc}

\usepackage{microtype}

\usepackage{inconsolata}

\usepackage{graphicx}
\usepackage{amsmath}
\usepackage{amssymb}
\title{AdaGReS:Adaptive Greedy Context Selection via Redundancy-Aware Scoring for Token-Budgeted RAG}
\author{
Chao Peng \quad Bin Wang \quad Zhilei Long \quad Jinfang Sheng \\
Central South University \\
Yizhi Intelligent (YZInt) \\
\texttt{chao.peng@yzint.cn}
}

\begin{document}
\maketitle
\begin{abstract}
Retrieval-augmented generation (RAG) is highly sensitive to the quality of selected context, yet standard top-k retrieval often returns redundant or near-duplicate chunks that waste token budget and degrade downstream generation. We present AdaGReS, a redundancy-aware context selection framework for token-budgeted RAG that optimizes a set-level objective combining query–chunk relevance and intra-set redundancy penalties. AdaGReS performs greedy selection under a token-budget constraint using marginal gains derived from the objective, and introduces a closed-form, instance-adaptive calibration of the relevance–redundancy trade-off parameter to eliminate manual tuning and to adapt to candidate-pool statistics and budget limits. We further provide a theoretical analysis showing that the proposed objective exhibits $\varepsilon $-approximate submodularity under practical embedding similarity conditions, yielding near-optimality guarantees for greedy selection. Experiments on open-domain question answering (Natural Questions) and a high-redundancy biomedical (drug) corpus demonstrate consistent improvements in redundancy control and context quality, translating to better end-to-end answer quality and robustness across settings.
\end{abstract}

\section{Introduction}

Retrieval-augmented generation (RAG), first introduced by Lewis et al. (2020)\cite{1}, has rapidly developed into a mainstream technique for enabling large language models (LLMs) to incorporate external knowledge and enhance performance on knowledge-intensive tasks. By integrating external documents or knowledge chunks with large models, RAG allows systems to dynamically access up-to-date, domain-specific information without frequent retraining, thereby improving access to up-to-date and domain-specific information without frequent retraining. Dense passage retrievers such as DPR proposed by Karpukhin et al. (2020)\cite{2}, the ColBERT model by Khattab and Zaharia (2020)\cite{3}, as well as later architectures like REALM\cite{4} and FiD\cite{1}, have further improved the retrieval, encoding, and fusion mechanisms of RAG in practical applications. Today, RAG is widely applied in open-domain question answering\cite{6}, scientific literature retrieval\cite{7}, healthcare, enterprise knowledge management, and other scenarios, becoming a key paradigm for enabling LLMs to efficiently leverage external knowledge.

Despite the significant advancements brought by RAG, particularly in enhancing knowledge timeliness, factual consistency, and task adaptability for LLMs, overall performance is still highly dependent on the quality of context chunks returned by the retrieval module. A persistent challenge is how to ensure that the retrieved results are not only highly relevant to the user’s query but also exhibit sufficient diversity in content. Numerous empirical studies have found that systems tend to return overlapping or near-duplicate chunks under top-k retrieval, especially when documents are chunked densely or the corpus is highly redundant\cite*{8}\cite{9}. Such redundancy not only wastes valuable context window (token budget) but can also obscure key information, limiting the model’s capacity for deep reasoning, comparative analysis, or multi-perspective synthesis, ultimately undermining factual accuracy and logical coherence.

For instance, in multi-hop question answering and multi-evidence reasoning tasks, if the retriever mainly returns paraphrased but essentially identical chunks, the model will struggle to acquire complete causal chains or diverse perspectives. This type of pseudo-relevance phenomenon has been shown to be an important contributor to hallucinations in RAG systems: when lacking sufficient heterogeneous evidence, the model may rely on internal priors and produce superficially coherent but externally unsupported and erroneous content\cite{10}.

To address fragment redundancy and hallucination, Maximal Marginal Relevance (MMR) and its variants have been widely adopted in existing RAG systems as well as in emerging frameworks such as GraphRAG\cite{11} and FreshLLM\cite{12}. By balancing relevance and diversity in the set of retrieved candidates, MMR can reduce redundancy and improve coverage. While effective in practice, these approaches still suffer from notable limitations: (1) their weight parameters are highly dependent on manual tuning and cannot dynamically adapt to the structure of different candidate pools or token budgets; (2) they only support local greedy selection, making it difficult to achieve set-level global optimality and potentially missing the best combination of chunks.

To systematically solve the issues of context redundancy, limited diversity, and cumbersome parameter tuning in RAG, this paper proposes and implements a novel context scoring and selection mechanism based on redundancy-awareness and fully adaptive weighting. Specifically, we design a set-level scoring function that not only measures the relevance between each candidate chunk and the query, but also explicitly penalizes redundancy among the selected fragments. The entire scoring process is mathematically modeled as the weighted difference between a relevance term and a redundancy term, with a tunable parameter $\beta$ controlling the trade-off between them. Building on this, we further propose a dynamic and adaptive $\beta$ adjustment strategy: by analyzing the average length, mean relevance, and redundancy distribution of the candidate pool, we derive a closed-form solution for the redundancy weight that adapts to different queries and budget constraints. This strategy provides a principled closed-form estimate of $\beta$, eliminating the need for manual parameter tuning or external heuristics. We also provide engineering implementations for instance-level $\beta$, as well as interfaces for fine-tuning on small validation sets and domain-specific customization, improving the method’s robustness and usability in real-world, variable scenarios.

To validate the theoretical foundation and practical effectiveness of the proposed approach, we conduct a rigorous theoretical analysis of the redundancy-aware adaptive selection framework. By proving the $\varepsilon $-approximate submodularity of the objective function, we establish approximation guarantees for greedy selection under $\varepsilon $-approximate submodularity. Our analysis further reveals how the adaptive $\beta$ mechanism dynamically suppresses excessive redundancy, enhances coverage, and prevents performance degradation, especially in complex data distributions or under tight budget constraints. Experiments demonstrate that the proposed method significantly outperforms traditional baselines on key metrics such as answer quality, coverage, and redundancy control, both in open-domain and biomedical knowledge retrieval tasks.

The main contributions of this work are as follows:

(1)We propose a redundancy-aware, fully adaptive context scoring and selection framework that systematically addresses key challenges such as context redundancy and limited diversity in RAG scenarios;

(2)We design a set-level relevance-redundancy joint scoring function and derive a closed-form adaptive solution to the $\beta$ parameter, enabling dynamic, instance-specific and budget-specific trade-off without manual tuning;

(3)We provide a theoretical analysis and proof of $\varepsilon$-approximate submodularity for our objective, offering theoretical guarantees for the near-global optimality of the greedy selection algorithm.

\section{Related Work}
\subsection{Retrieval-Augmented Generation and Context Selection}
Modern RAG systems typically employ dense or hybrid retrievers, such as DPR\cite{2}, ColBERT\cite{3}, or bi-encoder models, for initial passage retrieval, followed by subsequent ranking or selection modules to assemble the final context. This architecture achieves state-of-the-art results on benchmarks like Natural Questions and MS MARCO \cite{6,14}, but practical deployment in real-world domains reveals challenges in retrieval accuracy, context selection quality, and robustness to distributional shifts \cite{8,15,16}.

Recent studies have focused on deeper aspects of context selection and retrieval effectiveness. For example, Xu et al. (2025) developed a token-level framework showing that simply expanding retrieved context can mislead LLMs and reduce answer quality. Other works have explored integrating structured knowledge: KG²RAG\cite{9}and similar knowledge-graph-based retrieval systems\cite{10}improve factual grounding, but also raise new questions about chunk granularity and overlap. Advances such as HeteRAG (Yang et al., 2025) and modular retriever-generator architectures\cite{11,12}reflect a move toward decoupling retrieval and generation representations.

Nevertheless, a persistent challenge in RAG remains the redundancy and overlap among selected context chunks. Most traditional retrievers prioritize chunk–query relevance, and even diversity-aware rerankers such as Maximal Marginal Relevance (MMR) are typically applied with fixed tradeoff weights and myopic greedy decisions; as a result, the selected context can still contain repeated or overlapping information at the set level. This wastes token budget and can degrade language model performance, e.g., by amplifying hallucinations or reducing factual precision. Although redundancy-aware and diversity-promoting methods have recently appeared—such as attention-guided pruning (AttentionRAG, Fang et al., 2025) and dynamic chunk selection—many of these solutions still rely on heuristic or static parameters and require manual tuning; moreover, only a subset explicitly optimizes a set-level objective under strict token constraints, which can limit robustness and scalability in practice\cite{14}. Additionally, many methods fail to address token constraints and semantic variability in industrial-scale deployments\cite{15,16,17}.

To address these deficiencies, we propose a redundancy-aware scoring framework that unifies query relevance and intra-set redundancy into a principled, set-level objective. Our approach employs a greedy selection algorithm, with theoretical guarantees based on approximate submodularity, to construct high-coverage, low-redundancy context sets. Critically, we introduce a closed-form, fully adaptive calibration of the redundancy tradeoff parameter, allowing the system to automatically adjust according to candidate pool properties and token budget—removing manual tuning and ensuring robustness across domains and scales. Experiments demonstrate that our approach achieves better coverage, less redundancy, and improved answer quality versus prior baselines. This framework bridges the gap between real-world RAG constraints and optimal context selection, providing an effective and theoretically grounded solution for modern RAG systems.

\subsection{Retrieval-Augmented Generation and Context Selection}
Balancing relevance and diversity has long been a central objective in retrieval-based systems. The classical Maximal Marginal Relevance (MMR) framework addresses this by selecting items that maximize similarity to the query while minimizing similarity to previously selected elements, thereby reducing redundancy\cite{18}. This foundational idea inspired several extensions across information retrieval and text summarization. For instance, diversity-promoting retrieval methods strengthen the anti-redundancy component to cover multiple semantic clusters, but their performance heavily depends on manually tuned relevance–diversity coefficients\cite{19}. Determinantal Point Processes (DPPs) model subset diversity through determinant-based selection, providing strong theoretical properties but suffering from high computational cost as candidate pools scale\cite{20}. Submodular optimization approaches generalize MMR to set-level selection using predefined utility functions, yet often rely on fixed or validation-tuned parameters, making them less responsive to the redundancy structure of new candidate pools\cite{21}. Embedding clustering or centroid-based selection enhances semantic coverage but may sacrifice fine-grained relevance and overlook subtle but crucial information\cite{22}.

Across these MMR‑related approaches, several limitations consistently appear:  
(1) they rely on fixed or manually tuned tradeoff parameters,  
(2) they optimize selection locally rather than globally, and  
(3) they do not adapt to the characteristics of the candidate pool, such as varying redundancy levels or semantic density.  
Most importantly for retrieval‑augmented generation, these methods are not designed to account for strict token constraints in RAG database retrieval, where selecting too many redundant chunks directly wastes the available token budget and degrades downstream generation quality.

Our method Adaptive Greedy Context Selection via Redundancy-Aware Scoring addresses these limitations by introducing a fully adaptive redundancy-aware scoring function that calibrates the relevance–redundancy tradeoff according to candidate‑pool statistics and token budget, and integrates this improved scoring mechanism directly into the RAG context selection process, enabling efficient, redundancy‑controlled retrieval under realistic token constraints.

\subsection{Selection Algorithms and Theoretical Guarantees}
Selection algorithms grounded in submodular optimization have become foundational for data subset selection, document summarization, and retrieval-augmented generation. The concept of submodularity—first formalized in combinatorial optimization literature (Nemhauser et al., 1978)\cite{23}—describes set functions exhibiting the diminishing returns property: the marginal gain of adding an item to a smaller set is greater than adding it to a larger set. This property is critical because it enables efficient greedy algorithms to obtain strong theoretical guarantees. In particular, the seminal result proved that a simple greedy algorithm achieves at least a ($1-\frac{1}{e}$)-approximation for maximizing any monotone submodular function subject to a cardinality constraint. In token-budgeted RAG, the constraint is cost/budget-based rather than purely cardinality-based, but the submodularity framework remains valuable for motivating efficient greedy-style approximations under such constraints.

In context selection and related problems, submodular set functions have been widely adopted to model both relevance and diversity. For example, Lin and Bilmes (2011)\cite{24}leveraged submodular functions for extractive document summarization, demonstrating empirical and theoretical benefits in content coverage and redundancy reduction. More recently, Wei et al. (2015)\cite{25}and Mirzasoleiman et al. (2016)\cite{26}extended these ideas to large-scale data subset selection in machine learning pipelines, relying on submodularity to enable scalable and theoretically sound selection algorithms even as candidate pool size grows.

Our method, Adaptive Greedy Context Selection via Redundancy-Aware Scoring, inherits these theoretical properties: the redundancy-aware scoring function we propose exhibits approximate submodularity under realistic embedding distributions, allowing a greedy selection procedure to provide provable near-optimality for token-budgeted selection tasks in RAG. Detailed theoretical analysis and formal guarantees for our approach are presented in Section 4.

\section{Method}
\subsection{Redundancy-Aware Scoring Function}
Retrieval-augmented generation (RAG) pipelines often suffer from redundant or highly overlapping context selections, especially under strict token budgets. Conventional similarity-based retrievers maximize the relevance between the query and selected chunks but largely ignore information redundancy within the selected set, resulting in inefficient budget usage and repetitive evidence.

To address this, we introduce a redundancy-aware scoring function that jointly considers the relevance of each candidate chunk to the query and penalizes intra-set redundancy. Given a query embedding $q \in \mathbb{R}^d$ and a set of candidate chunk embeddings $\mathcal{V} = \{c_1, \ldots, c_N\}$,where q and all ci are L2-normalized to unit norm, we aim to select a subset $C \subset \mathcal{V}$ that maximizes the total query-aligned evidence mass (i.e., $\sum_{c \in C} \mathrm{sim}(q, c)$) while minimizing duplication.

Formally, our scoring function for any candidate subset $C$ is defined as:
\begin{equation}
  \label{eq:example}
  F(q, C) = \alpha S_{qC}(q, C) - \beta S_{CC}(C)
\end{equation}
where $S_{qC}(q, C) = \sum_{c \in C} \mathrm{sim}(q, c)$ measures the total relevance between the query and the selected chunks.$S_{CC}(C) = \sum_{i < j, c_i, c_j \in C} \mathrm{sim}(c_i, c_j)$ measures the total redundancy (pairwise similarity) among the selected chunks.Note that $S_{CC}(C)$ scales with the set size, so the appropriate magnitude of $\beta$ depends on the expected number of selected chunks under the token budget; we address this via an instance-adaptive closed-form $\beta$ in Section 3.3.Here, $\mathrm{sim}(a, b)$ denotes a non-negative cosine similarity, $\mathrm{sim}(a, b) = a^\top b$ under unit norm. The hyperparameters $\alpha > 0$ and $\beta \geq 0$ control the tradeoff between relevance and redundancy.When $\beta = 0$, the function reduces to standard relevance maximization, equivalent to vanilla similarity-based selection.As $\beta$ increases, the method penalizes redundant chunks more strongly, promoting diversity within the selected context.

Our approach generalizes the widely used maximal marginal relevance paradigm by making the relevance–redundancy tradeoff explicit in a set-level objective. While MMR is commonly implemented as a sequential greedy reranking rule with a fixed tradeoff weight, our formulation defines subset quality directly via $F(q,C)$, which provides a principled target for token-budgeted selection and motivates instance-adaptive calibration of $\beta$.

In summary, the proposed scoring function establishes a unified objective that maximizes coverage and information diversity, forming the basis for subsequent selection algorithms.

\subsection{Greedy Context Selection with Token Budget}
Maximizing the redundancy-aware objective $F(q, C)$ over all possible chunk subsets is a classical NP-hard combinatorial problem; exact search is infeasible for realistic candidate pool sizes. To efficiently approximate the optimal solution, we adopt a greedy selection algorithm that incrementally builds the context set by always choosing the candidate chunk with the highest marginal gain at each step, subject to the token budget constraint.

Formally, let $C_{\text{cur}}$ denote the current set of selected chunks. For any candidate $x \notin C_{\text{cur}}$, we define the marginal gain as:
\begin{equation}
  \label{eq:example}
  \begin{matrix}
  \Delta F(x \mid C_{\text{cur}})= F(q, C_{\text{cur}} \cup \{x\}) - F(q, C_{\text{cur}}) \\= \alpha\,\mathrm{sim}(q, x) - \beta \sum_{c \in C_{\text{cur}}} \mathrm{sim}(x, c)
  \end{matrix}
\end{equation}
At each iteration, we select
\begin{equation}
\label{eq:example}
x^* = \arg\max_{x \in \mathcal{V} \setminus C_{\text{cur}}} \Delta F(x \mid C_{\text{cur}})
\end{equation}
and add $x^*$ to $C_{\text{cur}}$, provided that adding $x^*$ does not exceed the overall token budget $T_{\max}$. The process stops when (1) no remaining chunk yields a positive marginal gain, or (2) the token limit would be exceeded by any further addition.

The complete greedy selection procedure is summarized below:
\begin{table}[h]
	\centering
	\renewcommand{\arraystretch}{1.2}
	\begin{tabular}{p{0.95\linewidth}}
		\hline
		\textbf{Algorithm 1: Greedy Context Selection with Token Budget} \\ \hline
		
		\textbf{Input:} query embedding $q$; candidate set $V=\{c_1,\ldots,c_N\}$; token budget $T_{\max}$; scoring weights $\alpha,\beta$; token length function $\ell(\cdot)$ \\
		
		\textbf{Output:} selected subset $C^*$ \\
		
		1. Initialize $C \leftarrow \emptyset$, $\text{total\_tokens} \leftarrow 0$ \\
		
		2. \textbf{while} $\text{total\_tokens} < T_{\max}$ \textbf{and } $\exists x \in V\setminus C$ with  $\Delta F(x|C)>0$ \textbf{do} \\
		\hspace{1em} Select $x^* \leftarrow \arg\max_{x \in V\setminus C} \Delta F(x|C)$ \\
		\hspace{1em} \textbf{if} $\text{total\_tokens} + \ell(x^*) \leq T_{\max}$ \textbf{then} \\
		\hspace{2em} $C \leftarrow C \cup \{x^*\}$ \\
		\hspace{2em} $\text{total\_tokens} \leftarrow \text{total\_tokens} + \ell(x^*)$ \\
		\hspace{1em} \textbf{else} continue \\
		
		3. \textbf{return} $C$ \\
		\hline
	\end{tabular}
\end{table}

\subsubsection{Limitation of Greedy Selection: Local Optima}

It is important to recognize that greedy selection, by its very nature, only optimizes the marginal gain at each step. This locally optimal decision process does not guarantee a globally optimal solution—especially in cases where initial non-optimal choices could lead to better overall combinations. In other words, the algorithm may become trapped in local optima and miss out on chunk sets that would yield higher objective scores if considered jointly.

\subsubsection{Theoretical Justification: Approximate Submodularity}

Despite this limitation, our design leverages important theoretical properties of the scoring function. As we will detail in Section 4, the redundancy-aware objective $F(q, C)$ exhibits $\epsilon$-approximate submodularity under typical embedding distributions, where chunk-to-chunk similarities are generally small. This property enables the greedy algorithm to achieve solutions that are provably close to the global optimum, with a bounded approximation gap. Our later analysis provides formal guarantees for the quality of greedy selection in this context.

In summary, greedy selection offers a practical and highly efficient approach to redundancy-aware context selection under token constraints. Its effectiveness is theoretically grounded by the approximate submodular nature of our scoring function, as will be established in the following sections.

\subsection{Adaptive $\beta$ for Dynamic Token Constraints}
The tradeoff parameter $\beta$ in the redundancy-aware scoring function plays a critical role in balancing query relevance and intra-set redundancy. However, its value is sensitive to both the candidate pool properties and the available token budget. A fixed $\beta$ may lead to excessive redundancy or overly aggressive pruning, resulting in suboptimal context selection. Therefore, we propose a principled approach to adapt $\beta$ per instance, using simple statistics of the candidate pool and the current token constraint.

\subsubsection{Motivation and Problem Statement}

When the token budget $T_{max}$ is tight, the number of selectable chunks is limited, and a larger $\beta$ is typically needed to prevent redundancy from consuming the scarce budget.When the budget is ample, or when the candidate pool is inherently diverse, a smaller $\beta$ often suffices to prioritize relevance.Our goal is to automatically calibrate $\beta$ such that, at the expected context set size dictated by the token budget, the marginal gain of adding a new chunk just reaches zero—balancing the contributions of relevance and redundancy at the decision boundary.

\subsubsection{Theoretical Derivation}

Let the average token length of the top candidate chunks be $\bar{L}$.Under token budget $T_max$, the expected number of selectable chunks is approximated as

\begin{equation}
  \label{eq:example}
  \bar{k} \approx \frac{T_{\max}}{\bar{L}}
\end{equation}
To encourage the greedy process to stop around $\bar{k}$ , we set the expected marginal gain at the boundary to be approximately zero:
\begin{equation}
  \label{eq:example}
  \Delta F(x|\bar{C}) = \alpha\,\mathbb{E}_{x}[\mathrm{sim}(q, x)] - \beta \sum_{c\in\bar{C}} \mathbb{E}[\mathrm{sim}(x, c)] \approx 0
\end{equation}

where expectations are taken over the top-N candidate pool, and $\bar{C}$ is a typical set of $\bar{k}-1$ selected chunks.
We approximate the redundancy increment by assuming that, for a typical boundary candidate xxx, the average similarity to each previously selected chunk is close to the average chunk–chunk similarity within the top-N pool. This yields:
\begin{equation}
  \label{eq:example}
  \sum_{c \in \bar{C}} \mathbb{E}[\mathrm{sim}(x, c)] \approx (\bar{k} - 1) \cdot \mathbb{E}_{x \neq y \sim \nu^{\mathrm{top}}} [\mathrm{sim}(x, y)].
\end{equation}

Solving for $\beta$ gives the adaptive calibration:
\begin{equation}
  \label{eq:example}
  \beta^* = \frac{\alpha\,\mathbb{E}_{x\sim \mathcal{V}^{top}}[\mathrm{sim}(q, x)]} {\frac{\bar{k}-1}{2} \cdot \mathbb{E}_{x\neq y \sim \mathcal{V}^{top}}[\mathrm{sim}(x, y)]}
\end{equation}

where $\varepsilon >0 $ is a small constant used for numerical stability (e.g., when the pool redundancy estimate is near zero).
$\mathcal{V}^{top}$ is the top-N candidate chunks most similar to the query,$\mathbb{E}_{x\sim \mathcal{V}^{top}}[\mathrm{sim}(q,x)]$ is the average query–chunk similarity,$\mathbb{E}_{x\neq y\sim \mathcal{V}^{top}}[\mathrm{sim}(x,y)]$ is the average pairwise similarity within the top-N candidate pool, which serves as a proxy for typical redundancy.

This closed-form solution provides a fully adaptive way to tune $\beta$ for any given query and candidate set.

\subsubsection{Why Use Expectation?}

The identity of the final selected chunk in the greedy process is not known a priori, and its specific relevance/redundancy cannot be determined before selection completes. Using empirical averages over the candidate pool provides a stable, low-variance estimate of “typical” relevance and redundancy near the selection boundary, enabling robust automatic calibration across a wide range of queries and corpus redundancy profiles.

\subsubsection{Practical Implementation}

To implement adaptive $\beta$ in practice:

1. Compute the average token length $\bar{L}$ of the top-N candidate chunks.

2. Estimate the expected set size $\bar{k} \approx T_{\max} / \bar{L}$ under the token budget.

3. Compute the average query–chunk similarity
$$\mathbb{E}_{x \sim \mathcal{V}^{top}} [\mathrm{sim}(q, x)].$$

4. Estimate the average pairwise redundancy within $\mathcal{V}^{top}$
$$\mathbb{E}_{x \neq y \sim \mathcal{V}^{top}} [\mathrm{sim}(x, y)]$$
	Since an exact computation is $O(N^2)$, we either use a modest $N$ or estimate it by sampling random pairs from $\mathcal{V}^{top}$.

5. Plug the statistics into the formula to compute $\beta^*$, and apply $\beta^*$ in greedy selection.

In practice, these computations add only lightweight overhead relative to embedding retrieval and scoring, and can be performed on a per-query basis.

Empirical Bias and Future Extensions

While $\beta^*$ is robust in most cases, highly skewed or noisy candidate pools may benefit from additional calibration. We therefore optionally introduce an empirical scaling/bias:
$$\beta = \lambda \cdot \beta^* + \beta_0$$
where $\lambda$ (scaling) and $\beta_0$ (bias) can be tuned on a small validation set or set to default values ($\lambda=1$, $\beta_0=0$) for fully automatic use. In engineering practice, $\beta$ can also be clipped to a reasonable range to avoid extreme pruning or extreme redundancy in outlier cases.

Moreover, these parameters could be learned via reinforcement learning or meta-learning to optimize downstream task performance or user preferences. We leave such learning-based calibration as future work.

In summary, adaptive $\beta$ is not merely an empirical trick. By calibrating $\beta$ using observed average redundancy and the budget-implied set size, we align the selection behavior with the redundancy profile of the candidate pool. This calibration helps keep the objective well-behaved for greedy optimization across datasets with different redundancy characteristics, thereby bridging practical robustness and theoretical soundness.

\section{Theoretical Analysis}

\subsection{Modularity and Supermodularity of Objective Components}

Our redundancy-aware objective function is composed of two main components—a relevance term that encourages alignment between the selected context and the query, and a redundancy term that penalizes overlap within the selected set. To analyze the theoretical properties of our objective, we first separately examine the modularity and (super-)modularity of each term.

\subsubsection{Relevance Term: Modularity}
The relevance component is given by:

\begin{equation}
  \label{eq:example}
  S_{qC}(q, C) = \sum_{c \in C} \mathrm{sim}(q, c)
\end{equation}
where $\mathrm{sim}(q, c)$ denotes cosine similarity. This function is modular, meaning that the contribution of each chunk $c$ to the overall score is independent of the other selected chunks. Adding a chunk $x$ to any subset $A$ yields a constant marginal gain:
\begin{equation}
  \label{eq:example}
  S_{qC}(q, A \cup \{x\}) - S_{qC}(q, A) = \mathrm{sim}(q, x)
\end{equation}
This is independent of $A$, confirming strict modularity. Modular functions are a special case of submodular functions. Moreover, $S_{qC}$ is monotone non-decreasing when $s(q,c)\ge0$ for all candidates, which holds under our non-negative similarity definition.

\subsubsection{Redundancy Term: Supermodularity}

The redundancy penalty is defined as:

\begin{equation}
  \label{eq:example}
  S_{CC}(C) = \sum_{i<j,\, c_i, c_j \in C} \mathrm{sim}(c_i, c_j)
\end{equation}

This measures the total pairwise similarity among selected chunks. The marginal increase when adding a chunk $x$ to a set $A$ is:

\begin{equation}
  \label{eq:example}
  S_{CC}(A \cup \{x\}) - S_{CC}(A) = \sum_{c \in A} \mathrm{sim}(x, c)
\end{equation}

Now consider two sets $A \subseteq B$, with $x \notin B$:

\begin{equation}
  \label{eq:example}
  S_{CC}(B \cup \{x\}) - S_{CC}(B) = \sum_{c \in B} \mathrm{sim}(x, c)
\end{equation}

Since $B$ contains all elements of $A$, we have:

\begin{equation}
  \label{eq:example}
  S_{CC}(B \cup \{x\}) - S_{CC}(B) \geq S_{CC}(A \cup \{x\}) - S_{CC}(A)
\end{equation}

because the sum in $B$ includes all terms from $A$, plus additional terms from $B \setminus A$ ,Since $s(\dot,\dot)\ge$ by construction, the marginal penalty increases with set size, establishing supermodularity This demonstrates that the redundancy term is supermodular, the marginal penalty for adding a chunk increases with the size of the set.

\subsubsection{Overall Structure: Modular Minus Supermodular}

Our full objective is

\begin{equation}
  \label{eq:example}
  F(q, C) = \alpha S_{qC}(q, C) - \beta S_{CC}(C)
\end{equation}

which is a modular term minus a supermodular term. This decomposition motivates a closer analysis of submodularity and the resulting greedy guarantees under token-budget constraints (see §4.2). While the relevance term always yields constant, context-independent gains, the redundancy term introduces a context-dependent penalty that grows superlinearly with the size and density of the set.

\subsection{Failure of Strict Submodularity}

While our redundancy-aware objective $F(q, C)$ elegantly balances query relevance and intra-set redundancy, its mathematical structure as a modular minus supermodular function means that, in general, it does not satisfy strict submodularity. This section provides a detailed analysis of why this is the case, and lays the groundwork for our subsequent notion of approximate submodularity.

\subsubsection{Definition and Intuition of Submodularity}

Recall that a set function $f: 2^V \to \mathbb{R}$ is submodular if, for any $A \subseteq B \subseteq V$ and $x \notin B$,

\begin{equation}
  \label{eq:example}
  f(A \cup \{x\}) - f(A) \geq f(B \cup \{x\}) - f(B)
\end{equation}
This property, known as diminishing marginal returns, is crucial for guaranteeing the near-optimality of greedy algorithms.

\subsubsection{Analysis of Our Objective}

Given
\begin{equation}
  \label{eq:example}
  F(q, C) = \alpha S_{qC}(q, C) - \beta S_{CC}(C)
\end{equation}

the marginal gain of adding chunk $x$ to set $C$ is

\begin{equation}
  \label{eq:example}
  \Delta F(x \mid C) = \alpha\,\mathrm{sim}(q, x) - \beta \sum_{c \in C} \mathrm{sim}(x, c)
\end{equation}

Consider two sets $A \subseteq B \subseteq V$ and $x \notin B$. The difference in marginal gains is

\begin{equation}
  \label{eq:example}
  \Delta F(x \mid A) - \Delta F(x \mid B) = \beta \sum_{c \in B \setminus A} \mathrm{sim}(x, c)
\end{equation}

If $\beta>0$ and similarities are non-negative, then $\Delta F(x|A)-\Delta F(x|B)\ge 0$, i.e., marginal gains decrease as the set grows, which is consistent with diminishing returns. However, with cosine similarity, some pairwise similarities may be negative, and the sum$\sum_{c\in B\setminus A}\mathrm{sim}(x,c)$ can become negative, yielding $\Delta F(x|A)<\Delta F(x|B)$ and violating strict submodularity.

\subsubsection{Counterexample}

For strict submodularity, we would require
\begin{equation}
  \label{eq:example}
  \Delta F(x \mid A) \geq \Delta F(x \mid B)
\end{equation}

But as shown above,
\begin{equation}
  \label{eq:example}
  \Delta F(x \mid A) - \Delta F(x \mid B) = \beta \sum_{c \in B \setminus A} \mathrm{sim}(x, c)
\end{equation}

If there exists any $c \in B \setminus A$ such that $\mathrm{sim}(x, c) > 0$ and $\beta > 0$, the right-hand side is negative, so the submodularity condition is violated.

\subsubsection{Visualization of Marginal Gain}

\begin{figure}[t]
  \includegraphics[width=1.0\linewidth]{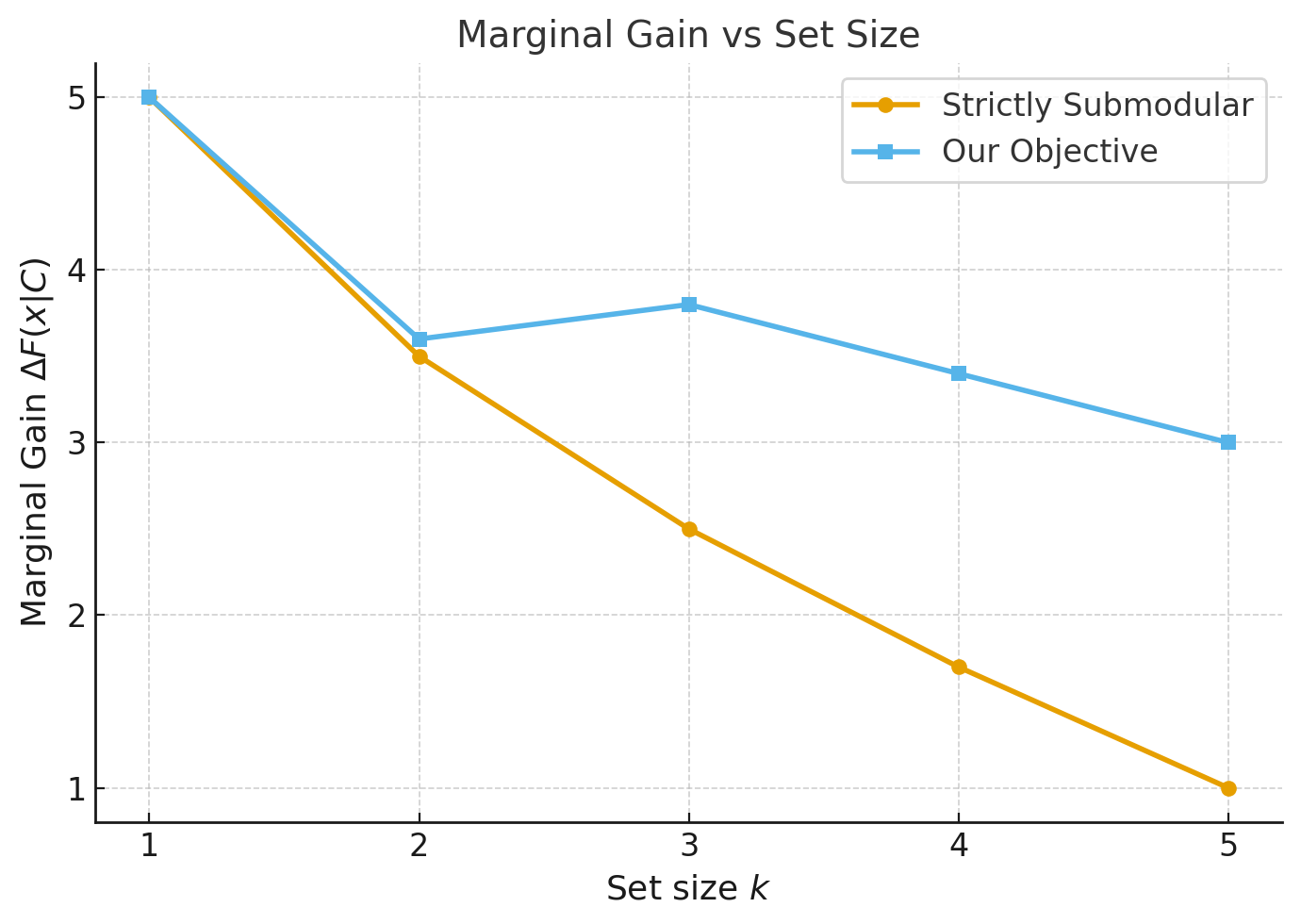}
  \caption {It illustrates the difference in marginal gain behaviors between strictly submodular objectives and our modular-minus-supermodular objective: strictly submodular objectives see marginal gains that consistently decrease as the set grows, while our modular-minus-supermodular objective may exhibit flat or even increasing marginal gains when high redundancy or non-overlapping candidates are present.}
\end{figure}

Interpretation

- The relevance term alone is modular and thus trivially submodular.
    
- The redundancy penalty grows with the set, causing marginal gain to decrease less than expected—or even to increase—when adding elements to larger sets.
    
- As a result, the overall function may exhibit supermodular effects, especially when $\beta$ or chunk similarities are large.

This limitation suggests that classical greedy guarantees for monotone submodular maximization under a cardinality constraint do not directly transfer to our setting. In addition, our selection is constrained by a token budget (knapsack-style) constraint, which further changes the conditions required for standard $1-\frac{1}{e}$-type results. In particular, greedy selection can fail to find the true global optimum, especially in worst-case settings with highly redundant candidate pools or poorly chosen $\beta$.

\subsubsection{Practical Implications}

Despite the lack of strict submodularity, in practical scenarios—such as typical embedding distributions where intra-chunk similarities are small—the objective may approximately satisfy diminishing returns, as we analyze in the next section. This justifies the strong empirical performance of greedy selection observed in our experiments.

\subsection{$\varepsilon$-Approximate Submodularity and Greedy Guarantee}

Given that our objective couples relevance with pairwise redundancy, its marginal gain depends on the already-selected set. In particular, when the similarity measure is allowed to take signed values, strict diminishing returns may not hold in the worst case. We therefore quantify how far our objective can deviate from submodularity via $\varepsilon$-approximate submodularity, and show that this deviation is bounded under mild similarity constraints—providing a principled justification for greedy selection in typical low-redundancy RAG candidate pools. we introduce and formalize the notion of $\epsilon$-approximate submodularity for our objective, show that—under typical conditions—this violation is tightly bounded, and explain how our adaptive $\beta$ strategy works hand-in-hand with theory to ensure practical near-optimality.

\subsubsection{Definition: $\varepsilon$-Approximate Submodularity}

A set function $f$ is said to be $\epsilon$-approximately submodular if for any $A \subseteq B \subseteq V$ and $x \notin B$,

\begin{equation}
  \label{eq:example}
  f(A \cup \{x\}) - f(A) \geq f(B \cup \{x\}) - f(B) - \epsilon
\end{equation}

where $\epsilon \geq 0$ quantifies the maximal deviation from strict submodularity. When $\epsilon$ is small, the function behaves nearly submodular, and greedy algorithms retain strong approximation guarantees.

\subsubsection{Bounding $\varepsilon$ for Our Objective}

Recall from the previous section that
\begin{equation}
  \label{eq:example}
  \Delta F(x \mid A) - \Delta F(x \mid B) = \beta \sum_{c \in B \setminus A} \mathrm{sim}(x, c)
\end{equation}

To bound this, suppose the pairwise cosine similarity between any two distinct candidate chunks is upper bounded by $\delta > 0$:

\begin{equation}
  \label{eq:example}
  \mathrm{sim}(x, c) \leq \delta \quad \forall\, x \neq c
\end{equation}

Suppose $|B \setminus A| \leq k$ (where $k$ is the maximal set size imposed by the token budget). Then,
\begin{equation}
  \label{eq:example}
  \Delta F(x \mid A) - \Delta F(x \mid B) \leq \beta k \delta
\end{equation}
Therefore, our objective is $\epsilon$-approximately submodular with
\begin{equation}
  \label{eq:example}
  \epsilon = \beta k \delta
\end{equation}

\subsubsection{Theoretical Role of Adaptive $\beta$}

It is important to note that the value of $\epsilon$—and thus the strength of our greedy approximation—depends directly on the choice of $\beta$. In scenarios where the candidate set exhibits high redundancy (i.e., large $\delta$ or large $k$), a fixed and overly large $\beta$ could result in a loose bound and poorer greedy performance.  
Our adaptive $\beta$ strategy (Section 3.3) is thus not merely an engineering choice, but is fundamentally grounded in theory: by estimating the average intra-candidate redundancy and expected set size for each instance, adaptive $\beta$ dynamically controls $\epsilon$ in the $\epsilon$-approximate submodularity bound. As redundancy or set size increases, $\beta$ is automatically reduced, constraining $\epsilon$ and preserving the near-optimality of the greedy solution.  
This coupling between parameter design and theoretical analysis is a key novelty of our approach.

\subsubsection{Greedy Guarantee under $\varepsilon$-Approximate Submodularity}

Prior work (e.g., Feige et al., 2011[28]; Horel $\&$ Singer, 2016[29]) shows that for $\epsilon$-approximately submodular functions, the standard greedy algorithm still provides strong guarantees:

\begin{equation}
  \label{eq:example}
  f(S_{\text{greedy}}) \geq \left(1 - \frac{1}{e}\right) \mathrm{OPT} - \frac{k \epsilon}{e}
\end{equation}

where $\mathrm{OPT}$ is the global optimum and $k$ is the maximal set size. Thus, the solution quality degrades only by an additive term proportional to $\epsilon$—which, with adaptive $\beta$, remains tightly bounded in practice.

\subsubsection{Practical Interpretation}

- In practice, chunk-to-chunk similarities are usually low due to semantic diversity in retrieved candidates, so $\delta$ is typically $\ll 1$.
    
- The penalty $\epsilon$ is therefore small, and greedy selection closely tracks the theoretical optimum.
    
- If candidate redundancy increases or $\beta$ is set too high, the additive gap grows, which can impact solution quality—but our adaptive $\beta$ mechanism is specifically designed to prevent this, by automatically calibrating $\beta$ based on candidate pool statistics (see Section 3.3).

\subsubsection{Summary and Theoretical Takeaway}

Our redundancy-aware objective exhibits $\epsilon$-approximate submodularity with
\begin{equation}
  \label{eq:example}
  \epsilon = \beta k \delta
\end{equation}

- Adaptive $\beta$ is essential for dynamically controlling $\epsilon$ and preserving greedy near-optimality, even under shifting candidate pool properties.
    
- Greedy selection remains nearly optimal in practical RAG scenarios.
    
- This analysis provides theoretical justification for both the strong empirical performance of our method and the use of greedy selection as a practical solution to redundancy-aware context selection.

\section{Experiments}
\subsection{Experimental Setup}
We conduct experiments on two distinct datasets to evaluate the generality and effectiveness of our method:

Domain-specific (Drug) Dataset:Our proprietary dataset consists of drug-related documents from the pharmaceutical domain, characterized by a high level of content redundancy.
Open-domain (Natural Questions) Dataset:We also evaluate our approach on the widely-used Natural Questions (NQ) dataset, following the standard practice of using the entire English Wikipedia as the retrieval corpus, with articles split into fixed-length chunks.

Retrieval and Selection:
For both datasets, we use the Conan-embedding-v1 model to generate dense embeddings for all chunks. Candidate passages are ranked and selected using either our redundancy-aware greedy selection (AdaGReS) or the similarity-only (top-k) baseline. For each redundancy penalty parameter $\beta$, we first run AdaGReS to determine the number of selected chunks k for each query, and then apply the similarity-only baseline with the same k, ensuring a fair and controlled comparison.

Evaluation Metrics:
Our primary evaluation metric is Intersection-over-Union (IOU), which measures the overlap between selected passages and the gold standard reference set. IOU in this context reflects both informativeness and precision: higher IOU indicates that the selected passages closely match the relevant ground-truth segments, whereas selecting excessive or irrelevant content will decrease IOU due to the expansion of the union set.

In addition, we conduct a qualitative human evaluation: for a set of representative queries, we compare answers generated by GLM-4.5-air using contexts retrieved by AdaGReS and the similarity-only baseline, respectively. This provides direct evidence of the end-to-end impact of different context selection strategies on QA performance.

Reproducibility:
All code, experiment scripts, and the domain-specific dataset will be released at [https://github.com/orderer0001/AdaGReS] and [https://huggingface.co/datasets/Sofun2009/yzint-drug-data] to ensure full transparency and reproducibility.

\subsection{Results and Analysis}
\subsubsection{Results on the Open Domain (NQ) Dataset}

\begin{figure*}[t]
  \includegraphics[width=1.0\linewidth]{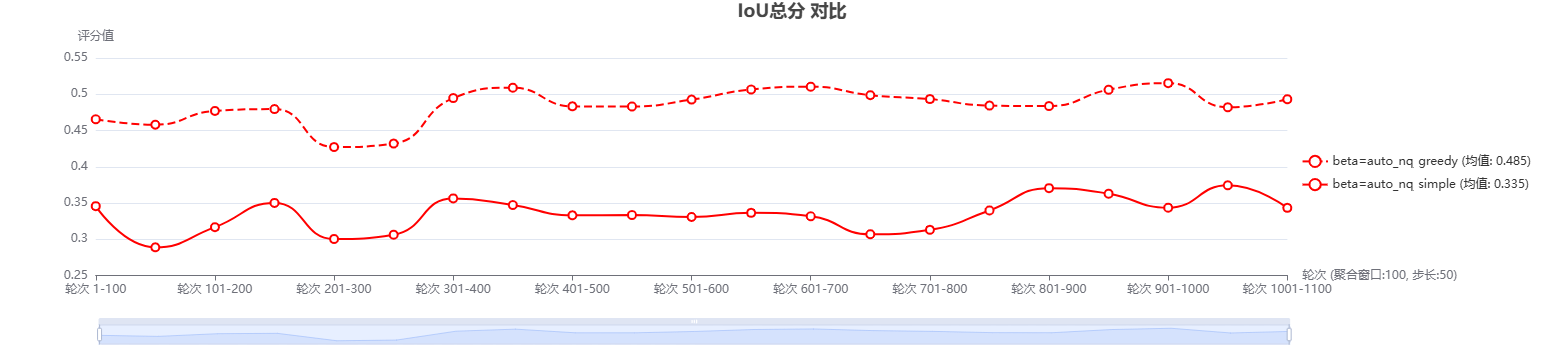}
  \caption {Visualization of IOU (Intersection over Union) scores between the dynamic $\beta$ method and the baseline method.}
\end{figure*}

\begin{figure*}[t]
  \includegraphics[width=1.0\linewidth]{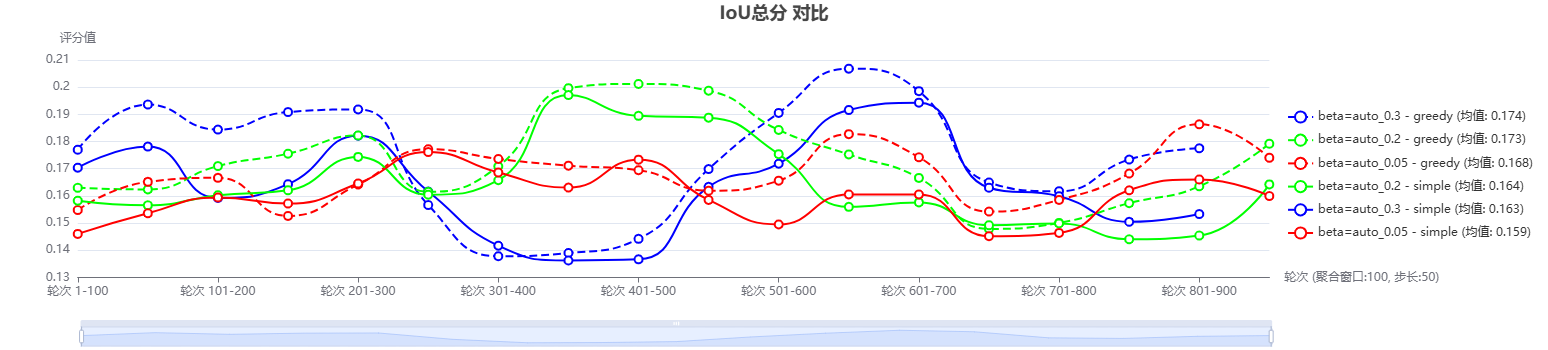}
  \caption {Visualization of Intersection over Union (IOU) Scores between the Dynamic $\beta$ Method and Baseline Methods under Different Redundancy Thresholds}
\end{figure*}

In open-domain question answering tasks (e.g., Natural Questions), answer information is often scattered across different semantic segments, which imposes higher requirements on the coverage breadth and anti-interference ability of context selection. Although the overall redundancy of candidate segments in this scenario is lower than that in specific domains, traditional retrieval methods still tend to encounter the issue of "high similarity but low information increment". Specifically, multiple high-scoring segments essentially only provide the same facts and fail to supplement effective new information for the question answering task.

In such scenarios, the AdaGReS model still demonstrates stable advantages, with its core lying in the built-in dynamic $\beta$ mechanism. This mechanism can adjust the intensity of redundancy penalty in real time based on the semantic distribution of the candidate pool: when there is less redundant information in the candidate pool, it moderately relaxes the diversity constraints to retain information relevance; when local semantics form dense clusters (i.e., similar segments appear in concentration), it automatically enhances deduplication capability to avoid repeatedly selecting similar segments. This dynamically balanced design enables the model to significantly reduce the redundancy rate while maintaining a high recall rate, and also effectively improves its support capability for multi-hop reasoning and multi-entity related questions.

From a quantitative perspective, Figure 2 (Visualization of Intersection over Union (IOU) Scores between the Dynamic $\beta$ Method and the Baseline Method in Open-Domain Question Answering Tasks) presents the performance comparison between the two methods. Experimental results show that the IOU score of the dynamic $\beta$ method is consistently higher than that of the baseline method across all experimental scenarios; meanwhile, the IOU score of the dynamic $\beta$ method remains stably at 0.15 and above throughout the entire testing process. The aforementioned results not only reflect the continuous performance advantage of the dynamic $\beta$ method over the baseline method, but also verify the stability of its IOU score, jointly supporting the advantages of the AdaGReS model in the accuracy and effectiveness of context selection from both intuitive and quantitative perspectives.
\begin{table*}[t]
  \centering
  \begin{tabular}{p{2cm}p{6.5cm}p{6.5cm}}
    \hline
    \textbf{Method} & \textbf{AdaGReS} & \textbf{Baseline} \\
    \hline
    Input 1 & 
    ``The MAF sensor detects the mass of air flowing into the engine and sends this data to the ECU\ldots'' &
    ``The MAF sensor detects the mass of air flowing into the engine and sends this data to the ECU\ldots'' \\
    
    Input 2 & 
    ``A mass airflow sensor (MAF) is used to determine the mass flow rate of air entering a fuel-injected internal combustion engine\ldots'' &
    ``A mass airflow sensor (MAF) is used to determine the mass flow rate of air entering a fuel-injected internal combustion engine\ldots'' \\
    
    Input 3 & 
    ``As a key detection component of the engine's air intake system\ldots'' &
    ``The MAF sensor determines the mass of air flowing into the engine's air intake system. The theory of operation of\ldots'' \\
    
    Response & 
    The mass air flow (MAF) sensor is a key component in the engine’s air intake system. It measures the mass flow rate of air entering a fuel-injected internal combustion engine and sends this data to the engine control unit (ECU). The ECU uses this information to accurately calculate the appropriate amount of fuel to inject, ensuring an optimal air-fuel mixture and efficient combustion. Additionally, the MAF sensor enables the ECU to adapt to environmental changes—such as variations in altitude and temperature—helping maintain consistent engine performance. A faulty MAF sensor can result in poor fuel economy, rough idling, or drivability issues. &
    The mass air flow (MAF) sensor measures the mass of air entering the engine and sends this information to the engine control unit (ECU). The ECU uses this data to calculate the appropriate amount of fuel to inject for optimal combustion. \\
    \hline
  \end{tabular}
  \caption{\label{tab:generation-comparison}
    Comparison of generated text between AdaGReS and the baseline model on MAF sensor description tasks.
  }
\end{table*}

\subsubsection{Quantitative Results on Domain-specific Dataset}

In specific domain knowledge retrieval tasks, considering that the employed embedding encoder has not been fine-tuned on the target domain dataset, its ability to distinguish subtle semantic differences within the domain is limited, which easily leads to highly semantically overlapping context fragments returned by retrieval. To alleviate the problem of information redundancy caused by this, we explicitly introduce a fixed penalty coefficient for redundant items themselves in the context scoring function of AdaGReS. That is, when calculating the comprehensive score of each candidate context, its redundancy metric is multiplied by a preset fixed weight less than 1, thereby directly weakening the contribution of redundant components to the final ranking. It should be emphasized that this penalty acts on the interior of redundant items rather than performing posterior weighting on the entire context score, thus enabling more refined regulation of the balance between redundancy and relevance.

As shown in Figure 3, we compared the Intersection over Union (IOU) scores of the AdaGReS method (with the introduced fixed redundancy penalty coefficient) and the baseline methods under different redundancy penalties (0.05, 0.2, 0.3). The experimental results show that even without domain fine-tuning, this strategy can bring consistent and stable performance improvements under all penalty settings. Although the overall gain is limited, its robustness verifies the effectiveness of this mechanism in high-redundancy professional scenarios.

The reasons for this limited improvement mainly come from two aspects: First, vertical domain knowledge itself has a high degree of semantic concentration. Contexts from different sources often revolve around the same core concepts, terms, or facts, resulting in natural content overlap and a relatively limited space for effective redundancy removal. Second, general pre-trained embedding models, in the absence of domain adaptation, struggle to accurately distinguish expressions that are semantically similar but informationally complementary (such as different descriptions of the same concept), leading to noise in the redundancy metrics themselves, which in turn limits the optimization upper limit that the fixed penalty mechanism can achieve.

To sum up, under the realistic constraint of not performing domain fine-tuning, by applying a fixed penalty coefficient to the redundant items themselves, AdaGReS can suppress the interference of semantic repetition on retrieval quality in a concise and interpretable manner. While maintaining computational efficiency, it effectively improves the diversity and information coverage of the results, providing a lightweight and robust redundancy control strategy for professional domain knowledge retrieval.

\subsubsection{Qualitative/Human Evaluation}

As shown in Table 1, AdaGReS focuses closely on the core requirement of “explaining the function of the mass air flow meter”. All returned segments are highly concentrated on functional descriptions without redundant repetition, and each output directly responds to the query, thus ensuring the relevance and conciseness of the results. In contrast, the baseline exhibits obvious redundancy issues in its retrieval results. It repeatedly outputs overlapping content related to the functions of the mass air flow meter, without performing effective screening and deduplication. These duplicated segments do not add new information to the answer of the query, but only lead to redundancy and bulkiness of the results.

The experimental results verify the design goals of AdaGReS: through the mechanisms of redundancy awareness, global optimization, and adaptive weighting, this method achieves synergistic improvement in context relevance, breadth of information coverage, and redundancy control capability. This optimization not only improves the utilization efficiency of token resources, but also enhances the knowledge integration and factual reasoning capabilities of generative models in complex retrieval-augmented scenarios.

\subsubsection{Ablation and Efficiency}
To further verify the role of each component, we conducted ablation experiments, replacing the adaptive $\beta$ with a fixed $\beta$ value.

\begin{figure*}[t]
  \includegraphics[width=1.0\linewidth]{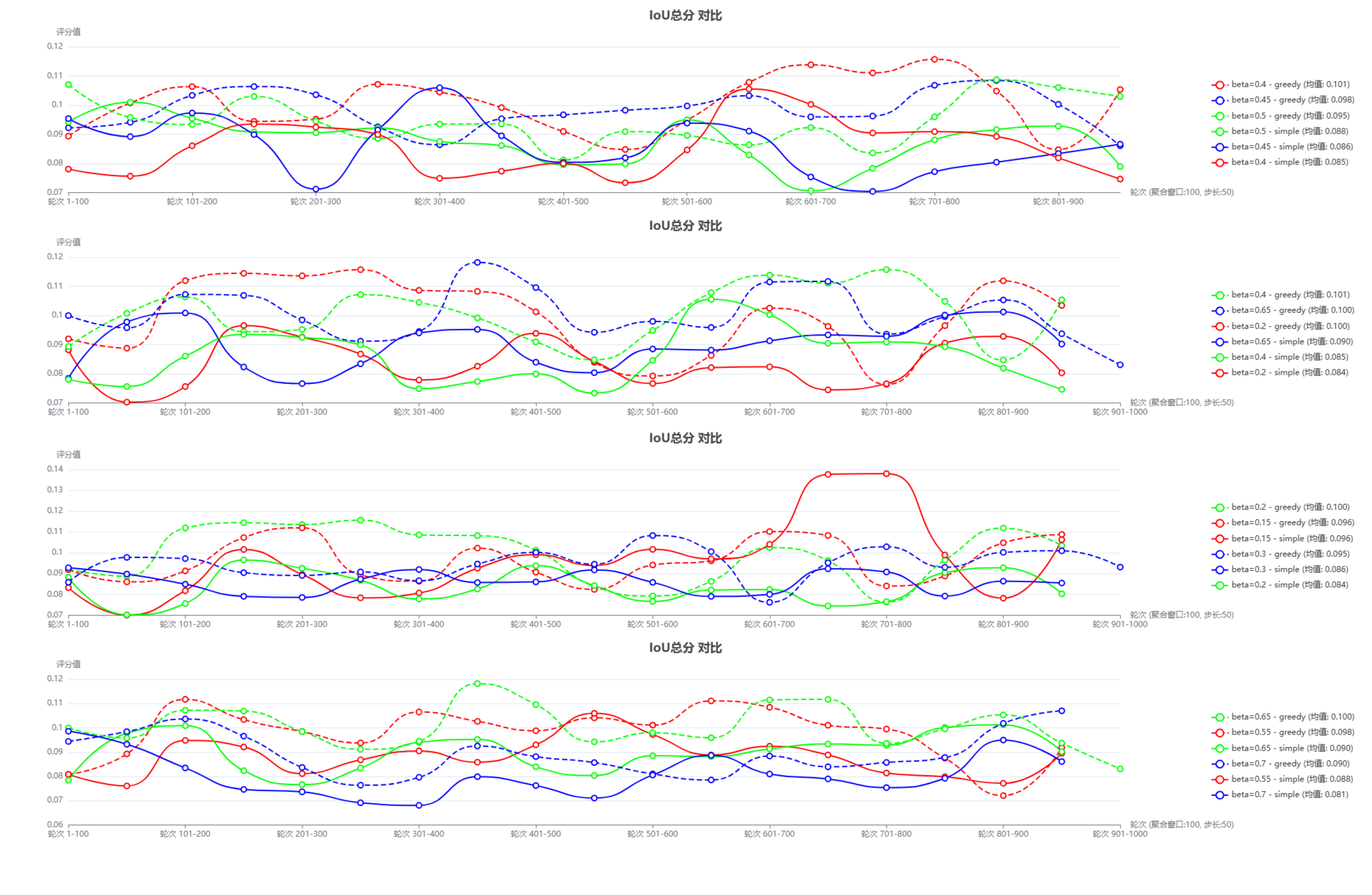}
  \caption {Comparison for $\beta$ = 0.55, 0.65, 0.7 (greedy vs. simple)}
\end{figure*}

Figures 4 show the IOU comparison between AdaGReS (greedy) and the similarity-only (simple) method under different redundancy penalty parameter $\beta$ settings. The dashed line represents AdaGReS, and the solid line represents the similarity-only method; the legend on the right lists the average IOU of each method under the corresponding $\beta$ value.

Specifically, for each $\beta$ setting, we first use AdaGReS to dynamically determine the number of selected blocks k, and then apply the similarity-only baseline method with the same k value to ensure fair comparison. In all figures, the dashed curve represents AdaGReS (greedy selection), and the solid curve represents the baseline method. In all test configurations, AdaGReS achieved the highest average IOU, fully verifying the robustness and general advantages of this method.

The experimental results show that in the pharmaceutical domain dataset, even when using a fixed $\beta$ value, the performance of AdaGReS is still better than the baseline method, although its IOU decreases slightly. This indicates that explicitly penalizing redundancy in the context selection process can effectively filter out more informative and diverse paragraphs, making them more consistent with the reference answers, further confirming the effectiveness and applicability of this mechanism.

We further evaluated the performance difference between AdaGReS and the similarity-only (top-k) selection method on the Natural Questions (NQ) dataset. NQ contains a large number of real user queries, and the retrieval corpus covers the full English Wikipedia, posing severe challenges to the information coverage and generalization capabilities of context selection algorithms in large-scale, open-domain scenarios.

Consistent with the domain-specific experiments, we used the Conan-embedding-v1 model to generate block-level embeddings, and for each query, we used AdaGReS and the baseline method to filter k paragraphs, where k is dynamically determined by AdaGReS under each $\beta$ parameter. The evaluation still uses Intersection over Union (IOU) as the main indicator.

\begin{figure*}[t]
  \includegraphics[width=1.0\linewidth]{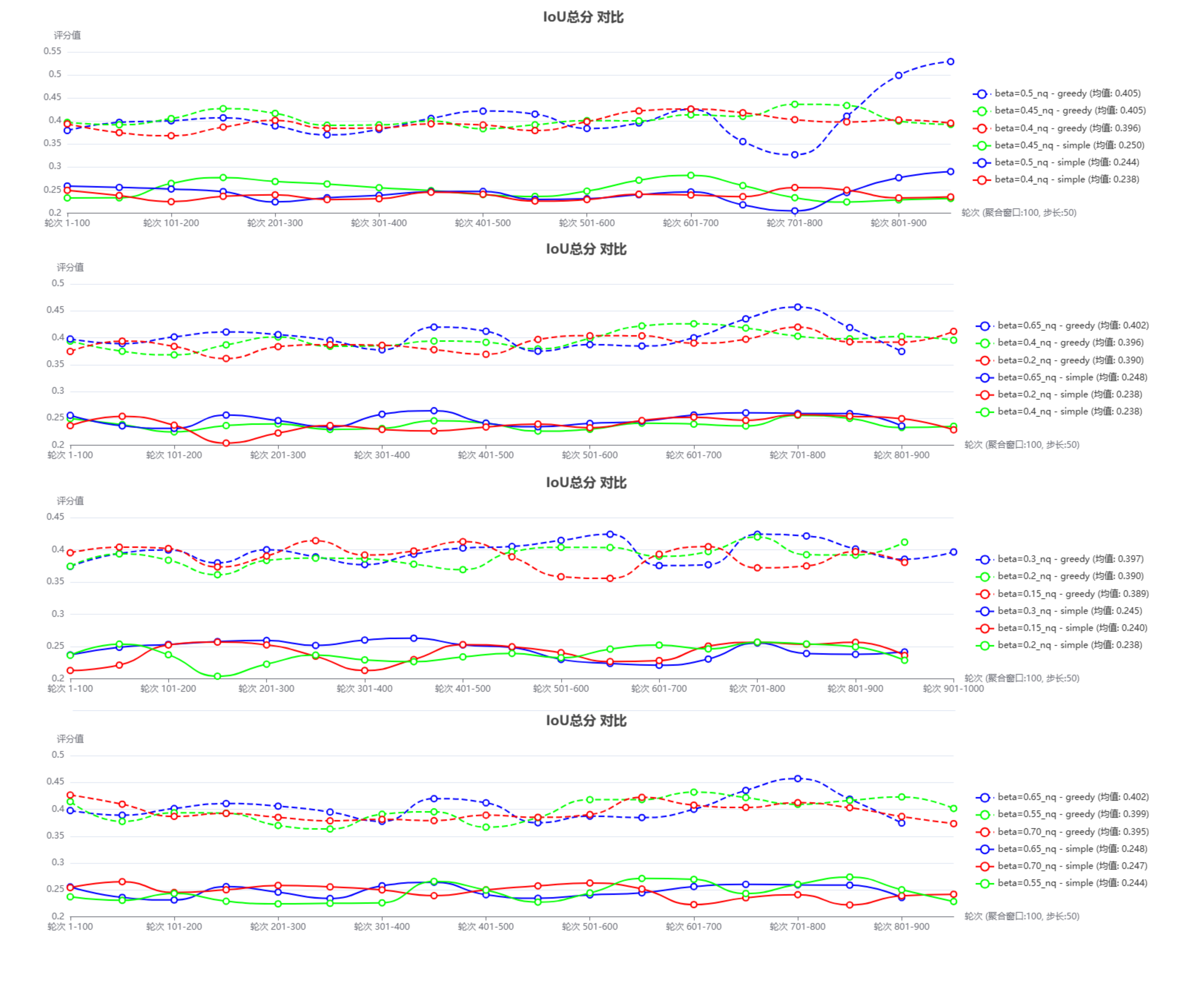}
  \caption {Comparison for $\beta$ = 0.55, 0.65, 0.7 (greedy vs. simple)}
\end{figure*}

Figures 5 show that AdaGReS also significantly outperforms the baseline method on the NQ task, especially in complex queries where answer information is scattered and multiple semantic perspectives need to be integrated, the IOU improvement is more prominent. This indicates that AdaGReS is not only applicable to highly redundant industry knowledge bases but also has excellent robustness and information integration capabilities in open-domain, large-scale corpus environments. For some ultra-long or extremely difficult-to-cover questions, the average IOU improvement of AdaGReS can reach 8–15 percentage points, further verifying its generality and practical value.

\subsection{Summary}
Overall, the experimental results fully validate the wide applicability and stable benefits of our proposed AdaGReS method in specific domains with highly redundant data (such as the pharmaceutical field) and complex retrieval environments in open domains (such as natural question datasets). Specifically, in terms of quantitative metrics: under different settings of the redundancy penalty parameter $\beta$, the IOU score of AdaGReS is consistently higher than that of the similarity-based baseline model. For complex open-domain queries, the IOU achieves a significant improvement. Even without domain fine-tuning, it can still achieve certain performance improvements in high-redundancy specific domain tasks. In terms of end-to-end generation effects: AdaGReS effectively avoids redundant repetitions in the retrieval context, enabling the generation model to produce more comprehensive, concise, and information-rich responses compared to the baseline model. In terms of human subjective evaluation: the retrieval results of AdaGReS are more focused on core query needs, without redundant overlapping content, and are significantly superior to the baseline model in terms of relevance and conciseness. In terms of computational efficiency and robustness: AdaGReS achieves refined redundancy control through an adaptive or fixed $\beta$ mechanism while maintaining high computational efficiency, and its advantages remain stable under various experimental configurations.

\section{conclusion}
In this work, we propose Adaptive Greedy Context Selection via Redundancy-Aware Scoring for Token-Budgeted RAG, a principled framework that enables efficient and globally optimized context selection for retrieval-augmented generation under strict token constraints. Our approach features a redundancy-aware set-level scoring function, combined with an adaptive mechanism that dynamically calibrates the relevance–redundancy tradeoff based on candidate pool statistics and token budget. This design effectively mitigates the long-standing challenges of context redundancy and information overlap, maximizing the informativeness and diversity of selected evidence without requiring manual parameter tuning. Empirical results and theoretical analysis demonstrate that our method consistently improves token efficiency and answer quality across various domains and scenarios.

In scenarios involving extremely long contexts with highly non-uniform redundancy distributions, our method may still present some minor limitations. When only a very small subset of candidate chunks is highly redundant while the rest are highly diverse, the greedy selection strategy, despite its theoretical guarantees, may occasionally fall short of achieving an ideal balance between redundancy suppression and coverage in practice. Nevertheless, such extreme distributions are uncommon in typical retrieval tasks, so this limitation does not affect the general applicability of the method. In the future, this aspect could be further improved through more refined diversity modeling or multi-pass selection strategies.


\bibliography{custom}

\appendix

\end{document}